\documentclass[10pt, a4paper]{article}

\usepackage[]{lrec-coling2024} 

        \hypersetup{
           breaklinks=true,   
           colorlinks=true,   
           pdfusetitle=true,  
        }

\title{Impoverished Language Technology: On the State of Class in NLP}
\title{Impoverished Language Technology: \\The Lack of (Social) Class in NLP}

\name{Amanda Cercas Curry, Zeerak Talat, Dirk Hovy} 

\address{Bocconi University, Mohamed Bin Zayed University of Artificial Intelligence, Bocconi University\\
         amanda.cercas@unibocconi.it, z@zeerak.org, dirk.hovy@unibocconi.it\\
         }

\abstract{
Since \citepos{labov1964social}
foundational 
work on the social stratification of language, linguistics has dedicated concerted efforts towards understanding the relationships between socio-demographic factors and language production and perception.
Despite the large body of evidence identifying significant relationships between socio-demographic factors and language production, relatively few of these factors have been investigated in the context of NLP technology. While age and gender are well covered, Labov's initial target, socio-economic class, is largely absent. We survey the existing Natural Language Processing (NLP) literature and find that only $20$ papers even mention socio-economic status.
However, the majority of those papers do not engage with class beyond collecting information of annotator-demographics.
Given this research lacuna, we provide a definition of class that can be operationalised by NLP researchers, and argue for including socio-economic class in future language technologies.
 \\ \newline  }

\begin{document}

\maketitleabstract

\section{Introduction}
A salient aspect of identity formation is the creation of class, or socio-economic identity.
Certain accents, phrases, and constructions are considered either upper, middle, or lower class, or a host of other references to socio-economic status.

\newcite{labov1964social} was the first to systematically investigate this relation. He observed that New Yorkers with higher socio-economic status tended to pronounced the /R/ sound after vowels, whereas the traditional dialect dropped it. 
He devised a study to quantify this observation by asking clerks in various department stores (a proxy for socio-economic status) for items found on the fourth floor, then recorded how many of Rs were dropped. 
He found a clear anti-correlation between the socio-economic status of the store (and hence presumably the speakers there) and the number of dropped Rs: the higher the status, the fewer dropped Rs.

Since \citeauthor{labov1964social}'s \citeyear{labov1964social} intervention on social stratification and language, linguistics has dedicated concerted efforts towards understanding how different socio-demographic factors influence the production and perception of language, and how speakers use them to create identity \cite{eckert2012three}.
Despite the large body of evidence showing the relationships between language and demographic factors (the ``first wave'' of socio-linguistic variation studies), relatively few socio-demographic factors have been investigated in the context of natural language processing (NLP) technology.

Existing work on socio-demographic factors has predominantly focused on how much a certain linguistic variable signals age, ethnicity, regional origin, and gender \cite{johannsen-etal-2015-cross}. 
Almost none of the NLP works have engaged with the second and third wave of sociolinguistics, i.e., how variation creates local identity and drives language change. To address this gap, we focus on socio-economic status in NLP.

Filling this lacuna could provide useful future research avenues for computational sociolinguistics and social science, which often depends on the proper stratification of data into socio-demographic categories.
More broadly, excluding a crucial sociodemographic factor like social class from consideration impoverishes NLP's capability to counteract social biases in its tools and datasets.


\paragraph{Contributions}
We document the lack of NLP work dealing with socio-economic status. We survey how socio-economic status is measured in the NLP literature and contrast this with metrics used in the social sciences. We conclude with some recommendations for future research.

\section{What is Social Class?}
 Social stratification refers to the grouping of people according to socio-economic status (SES) based on factors like income, education, wealth, and other characteristics, with different groups being distinguished in terms of power and prestige~\cite{saunders2006social}. 
 There are different systems of social stratification, including the Indian caste system, clans or tribes, and the Western hierarchical class system. 

Exactly how many social strata there are is unknown and likely varies from region to region.
However, in Western cultures it is common to see at least three strata referring to upper, middle and lower class people. 
Other systems refer to blue collar and white-collar jobs.  
Recently, the Great British Class Survey \citep[GBCS,][]{savage2013new} has taken an empirical approach to understanding the different social strata, and they propose a seven-level system for the United Kingdom. 
Their stratification is based on economic, social and cultural capital: elite, established middle class, technical middle class, new affluent workers, traditional working class, emergent service workers, and precariat. 
The derive these classes from a survey conducted over British citizens that received more than $160$K responses.

\section{The Impact of Social Class on Language}
Social class shapes people’s everyday experiences by granting or limiting access to resources. 
Social stratification has a significant impact on people beyond power and prestige, for instance, lower socio-economic status has been linked to worse health outcomes and higher mortality \cite{Saydah_Imperatore_Beckles_2013}. 
Socio-economic status also influences language: \citeauthor{labov1964social}'s germinal work showed there are differences in pronunciation, \newcite{bernstein1960language} showed that children from working class families had significantly smaller vocabularies even when general IQ was controlled for, and \newcite{flekova-etal-2016-exploring} has recently shown that there is significant lexical and stylistic variation between social strata. 

Socio-economic status affects language use from the very early stages of development.
\newcite{bernstein1960language} posits that language takes on a different role in middle- and working-class families, where middle-class parents encourage language learning to describe more abstract thinking. 
In working-class families, parents are \emph{limited} to more concrete and descriptive concepts. 
Parents from lower SES tend to interact less with their children, with fewer open-ended questions than parents from higher SES, which shapes language development \cite{clark2015first}. 
While \newcite{usategui1992sociolinguistica} shows that education reduces this language gap, one's level of education has traditionally been one of the key factors in determining social class and potential for upward social mobility. Moreover, a person's accent is a strong marker of SES, reported to lead to anxiety and discrimination \cite{sutton2022accent}.

Given the well-documented effects that socio-economic status has on language development and use, it stands to reason that NLP should carefully consider social class as a variable.

\section{Measuring socio-economic status}\label{sec:measuring_class}

As social class encompasses more than a single factor, e.g., income, measuring and classifying people according to socio-economic status is a non-trivial task. 
Social class may be measured objectively through measures of socio-economic status (SES) or subjectively, by asking participants to self-report.\looseness=-1

\paragraph{Objective} 
In terms of objective measures of social class, education, income, and occupation are the most widely used factors to measure social class \cite{kraus2012road}.  
Education affords access to higher salaries, more prestigious occupations and higher levels of cultural capital. 
Income is the most direct measure of an individual’s access to material goods and services. 
Finally, occupation is a strong indicator of prestige and other formative experiences.

More recently, the GBCS \cite{savage2013new} asked participants about their economic, social and cultural capital based on a framework described by \newcite{bourdieu2018distinction}. Economic capital refers to one's income and assets\footnote{Namely, whether one owns their house or rents it and the amount of savings.}, \emph{social capital} is measured in terms of the prestige of those in one's social circle\footnote{As determined by their status scores according to  the Cambridge Social Interaction and Stratification (CAMSIS) scale.} and \emph{cultural capital} in terms of the type of cultural activities in which they participate (e.g. attending the theatre vs football matches). 
Based on the results from the survey, \newcite{savage2013new} propose a seven-level stratification extracted from latent class analysis on 160K participants from the U.K. This fine-grained classification shows societal changes and fragmentation in the middle-class.

\paragraph{Subjective} In terms of subjective measurements, social class is how much one believes they have relative to others. 
People’s perception of where they stand in terms of social class has important psychological effect even when controlling for objective measures, supporting the idea that subjective class is an important measure. The general recommendation is to use the Macarthur scale \cite{adler2000relationship}, where people are asked to place themselves on a ladder, with higher levels representing those who are more privileged.

Because there may be discrepancies between one's subjective and objective SES, the American Psychological Association (APA)\footnote{\href{https://www.apa.org/pi/ses/resources/class/measuring-status}{APA: Measuring SES}} recommends measuring a participant's level of education, income, occupation, and family size and relationships, as well as subjective social status \cite{diemer2013best}. 
These recommendations add a layer of difficulty for researchers because (1) the questions are intrusive (for example in many countries it is considered impolite to discuss salaries) and refer to sensitive topics, and (2) while gender and race are a single data point, accurately measuring socio-economic status requires a minimum of four questions that need to be aggregated.

For these reasons, the majority of NLP work has focused on one particular aspect of social class as a proxy. For example, \newcite{lampos-etal-2014-extracting} use occupation only as a proxy and \newcite{flekova-etal-2016-exploring} use only income. However, income is only one small part of social class and does not capture its nuances and the over-emphasis on income and occupation of most social classification systems has been criticised by some feminist scholars for how it interacts with some social norms, e.g., the expectation of women to be homemakers \cite{skeggs1998formations, crompton2008class}.

\section{Survey of NLP Literature on Social Class}
Here, we analyse existing literature in NLP that deal with SES in some way. 
As a first step, we collect the bibliography file for the ACL anthology and search for the occurrence of terms in the titles and abstracts.\footnote{The full list of our search terms: `social class', `caste', and `socio-economic', `income', `education', `occupation', `white/blue collar', `upper/middle/lower class'} 
We also experimented with terms like `occupation' and `education' which may be used a proxy of social class, however we find that these papers focus on either gender bias (in the case of occupation), or education in and of itself without engaging with social class in any way. 
In addition, we refer back to feminist scholars who criticise metrics such as income and occupation as individual class markers in patriarchal societies (see \newcite{skeggs1998formations, crompton2008class}). 
Our initial search yielded 78 papers, however, many of them do not engage with the topic at all and only mention social class in passing, e.g. `communication between people from different religion, caste, creed, cultural and psychological backgrounds has become more direct'. 
After removing such papers, only 20 papers remain. 

Of these 
21 papers, four focus on NLP in low- and middle-income countries/regions for social equity but do not directly model language. 
Three papers collect SES-related metrics, but do not use this information in their analyses.
Finally, the remaining papers deal directly with modelling language according to SES (e.g. predicting the income of Twitter users).


Although (as expected) the majority of papers are dealing with English, we find a wide variety of languages given the small sample (English, Danish, French, Hindi, Marathi, Russian). 

\subsection{How NLP Measures SES}
So far, no systemic method to measure SES has been proposed for NLP.
Table \ref{tab:ses_measurements} shows the full list of papers and their SES measurement. 
The majority focus on one or two aspects of SES, such as income (9) and education (5). Only five papers refer explicitly to a class system, using a two- or three-level classification, however none use objective measures of class. One paper uses restaurant prices as a proxy for socio-economic status. In addition, in terms of granularity, many of these studies do not collect data on an individual level but rather use reported statistics for a given country/area using census data.  

We conclude from the survey that socio-economic status is rarely reported on in NLP literature, where most data is collected from urban citizens and university students, or from middle- to upper-class sources like news outlets. Low-SES is only specifically collected to cover this subset of the population for a given study, possibly due to the increased difficulty in accessing people from less privileged backgrounds when research is done in universities and affluent urban areas. However, sourcing data from lower-SES participants still affords quality data while also offering supplemental income \cite{abraham-etal-2020-crowdsourcing}. As NLP technologies are becoming increasingly ubiquitous in society, we should endeavour to include all sub-populations to ensure fair and equitable technologies. In addition, NLP also serves a role as cultural anthropology and should reflect the reality of language use across populations.

\begin{table*}[h]
    \centering
    \begin{tabular}{l|c| c} 
          & \textbf{Measurement} & \textbf{Granularity} \\ \hline
         \newcite{lampos-etal-2014-extracting} & Unemployment & Country\\
         \newcite{preotiuc-pietro-etal-2015-analysis} & Occupation & Individual \\
         \newcite{flekova-etal-2016-exploring} & Income & Individial\\
         \newcite{hasanuzzaman-etal-2017-temporal} & Income & Individual \\
         \newcite{giorgi-etal-2018-remarkable} & Income, Education & County (census data)\\
         \newcite{zamani-etal-2018-residualized} & Income, Education, Unemployment & Country-wise\\
         \newcite{degaetano-ortlieb-2018-stylistic} & Class (high, low) & Individual\\
         \newcite{van-etal-2019-language} & Income, poverty education & State-level (census) \\
         \newcite{jawahar-seddah-2019-contextualized} & Income, geolocation & Neighbourhood-level\\
         \newcite{basile-etal-2019-write} & Restaurant price & Individual \\
         \newcite{ghazouani-etal-2019-assessing} & socio-economic status & Mixed\\
         \newcite{abraham-etal-2020-crowdsourcing} & Income, area & Group \\
         \newcite{tafreshi-etal-2021-wassa} & Income, education & Individual \\
         \newcite{abbasi-etal-2021-constructing} & Income, education & Individual\\
         \newcite{stromberg-derczynski-etal-2021-danish} & SES (high, mix, unknown) & Aggregated by dataset\\
         \newcite{van-boven-etal-2022-intersection} & Low-income countries & Country \\
         \newcite{ngao-etal-2022-detecting} & Low-income countries & Country\\
         \newcite{grutzner-zahn-rehm-2022-introducing} & GDP & Country\\
        \newcite{cole-2022-crowdsourced} & Class (high, low) & Individual \\
         \newcite{malik-etal-2022-socially} & Caste, occupation & General (bias)\\
         \newcite{hrzica-etal-2022-morphological} & Class (middle) & Group \\
    \end{tabular}
    \caption{
    Papers included in the survey along with the metric used to assess socio-economic status and 
    the granularity of the metric is.}
    \label{tab:ses_measurements}
\end{table*}

\section{Measuring Class for NLP Practitioners}
A standout finding from the survey is the lack of a unified measurement of socio-economic status in the literature. This makes it difficult to compare language varieties across studies and datasets or to get a clear picture of whose language exactly NLP research is mapping. 
In addition, it remains unclear how some of the proxy metrics used (such as geolocation or income) relate to class and language variety. 
Section \ref{sec:measuring_class} provided an overview of possible metrics for socio-economic status. Based on this, we make some recommendations for the NLP community:
\begin{itemize}
    \item Where participants are explicitly recruited for a study (i.e. the data is not collected en masse from social media), researchers should try to take objective measures of the participants' SES (see \ref{sec:measuring_class}).
    
    \item Alternatively, aim to report at least their subjective SES\footnote{As this is a one-point measure and is not as intrusive as concrete questions about income.} following the Macarthur scale, which has already been scientifically validated.
    
    \item Data collected from social media rarely contains such information at the individual level -- while people may list their occupation in their biographies, these tend to be biased towards high-prestige occupations \cite{guo2024biographies}, leaving lower status occupations in the dark. Researchers should endeavour to contextualise the data collected by considering and appropriately describing the general statistics of the social media platform used (e.g. \newcite{ghazouani-etal-2019-assessing} aim to asses the socioeconomic status of X -- formerly known as Twitter -- users).
    
    \item If all else fails, report socio-economic status in whichever way is possible (for example by using some of the proxy metrics discussed in this paper, or e.g.~\newcite{cercas-curry-etal-2021-convabuse} use level of education and the university's prestige, though these may be dependent on culture).
\end{itemize}

Properly documented and contextualised data collection is crucial for more equitable NLP (and more broadly linguistics) research.

\section{Related Work}
While there has been an uptick in the number of papers tackling gender and racial bias in NLP, work considering other under-privileged communities has lagged behind. In a survey of Computational Sociolinguistics, \citet{nguyen-etal-2016-survey} point to the lack of self-reported explicit labels in online user profiles as possible cause, with work focusing on occupation (often mapped to income or other variables. 

As a way to mitigate and document biases in NLP, \newcite{bender2018data} ask for the socio-economic status of both the speakers and the annotators to be declared, however they do not suggest any standardised way to measure or report this.
\newcite{field-etal-2021-survey} (from whom we borrow our methodology) conduct a survey focus on race but also call for more diversity in NLP in terms of the broader inclusion other underprivileged people such as those from lower socio-economic status.

\section{Directions for Future Research}
NLP has two main purposes: (1) as a descriptive tool for current language use, and (2) as a service for everyday life. In both cases, NLP must be able to serve and represent people equally. With the ubiquity of language technologies, it is no longer the case that only the wealthiest need to access these but by excluding those less privileged from NLP datasets we are crippling our technologies in their ability to serve all of humanity. NLP systems are enforcing a standard of language by limiting the lects they represent. 

Future research should establish how the current and proposed metrics and models represent language varieties. Some questions for future work are whether subjective class is more predictive of language use than objective class, how fine-grained of a classification is needed, and how socio-economic status interacts with other socio-demographic factors like age, gender, race or region when it comes to language. 

Creating new datasets and tools to identify social class distinctions in text would not only help build fairer NLP technology, but also benefit related disciplines that use NLP tools to stratify their data along socio-demographic lines.

\section{Conclusion}
We have explored the definition of social class, how it can be measured and its effect on language use. We then surveyed existing work in the ACL anthology dealing with socio-economic status and its components and found significant gaps. First, very little work considers social class despite its well-documented effect on language, and second, there is currently no systematic way to measure socioeconomic status in the few papers that do report it. We encourage researchers to ensure diversity in collected datasets and to be more diligent in reporting accurate socio-demographics for their participants. 

\section*{Limitations}
Our survey has provided an overview of work in NLP engaging with socioeconomic status, however, we focused only on papers included in the ACL anthology. This means some work that has been published in other venues has been excluded. In addition, our keyword search is done only on titles and abstracts, though any work meaningfully engaging with SES is likely to mention it. 
Finally, our survey and suggestions have focused mainly on Western definitions of SES and social class that may not be applicable in other cultures.

\section*{Acknowledgements}
Amanda Cercas Curry and Dirk Hovy were supported by the European Research Council (ERC) under the European Union’s Horizon 2020 research and innovation program (grant agreement No.\ 949944, INTEGRATOR). They are members of the MilaNLP group and the Data and Marketing Insights Unit of the Bocconi Institute for Data Science and Analysis.

\section{Bibliographical References}\label{sec:reference}

\bibliographystyle{lrec-coling2024-natbib}
\bibliography{anthology,custom}

\end{document}